%% file: main.tex
\documentclass[11pt,a4paper]{article}
\usepackage[hyperref]{naaclhlt2019}
\usepackage{times}
\usepackage{latexsym}

\aclfinalcopy
\usepackage{color}
\usepackage{amsfonts}
\usepackage{graphicx}
\usepackage{amsmath,amsfonts,amssymb,bm}
\usepackage{booktabs} 
\usepackage{array}
\usepackage{boxedminipage}
\usepackage{paralist}
\usepackage{url}
\usepackage{diagbox}
\usepackage{enumitem}
\usepackage{paralist}
\definecolor{darkspringgreen}{rgb}{0.05, 0.5, 0.06}
\usepackage{array}
\usepackage{pifont}

\usepackage{microtype}      % microtypography
\usepackage{comment}        % bulk comment
\usepackage{amsopn}
\usepackage{xcolor}
\usepackage{multirow}
\usepackage{subcaption}
\usepackage{tikz,pgfplots}

\usepackage{tabularx}
\usepackage{colortbl}
\usepackage{hhline}

\newcommand{\red}[1]{\textcolor{red}{#1}}

\usepackage{url}

\newcommand{\blue}[1]{\textcolor{blue}{#1}}

\newcommand{\sys}{{\sc DyGIE}}

\newcommand{\cmark}{\ding{51}}%
\newcommand{\xmark}{\ding{55}}%

\newcolumntype{L}[1]{>{\centering\let\newline\\\arraybackslash\hspace{0pt}}m{#1}}
\newcolumntype{C}[1]{>{\centering\let\newline\\\arraybackslash\hspace{0pt}}m{#1}}
\newcolumntype{R}[1]{>{\centering\let\newline\\\arraybackslash\hspace{0pt}}m{#1}}

\newcommand{\luanyi}[1]{\textcolor{red}{[#1 - YL]}}

\newcommand{\PreserveBackslash}[1]{\let\temp=\\#1\let\\=\temp}
\newcolumntype{C}[1]{>{\PreserveBackslash\centering}p{#1}}
\newcolumntype{R}[1]{>{\PreserveBackslash\raggedleft}p{#1}}
\newcolumntype{L}[1]{>{\PreserveBackslash\raggedright}p{#1}}

\title{A General Framework for Information Extraction \\  using Dynamic Span Graphs}
\author{
  Yi Luan$^\dag$ \quad Dave Wadden$^\dag$ \quad Luheng He$^\ddag$ \quad Amy Shah$^\dag$ \quad Mari Ostendorf$^\dag$ \quad Hannaneh Hajishirzi$^{\dag*}$ \\
$^\dag$University of Washington\\
 $^{*}$Allen Institute for Artificial Intelligence\\
    $^\ddag$Google AI Language\\
  {\{luanyi, dwadden, amyshah, ostendor, hannaneh\}@uw.edu}\\
  luheng@google.com\\
}

\date{}

\begin{document}

\maketitle

\newcommand\luheng[1]{[\textcolor{orange}{LH: {#1}}]}

\input{01-Abstract.tex}

\input{02-Introduction.tex}

\input{03-RelatedWork.tex}

\input{04-Model.tex}
%   \input{04-DWModel.tex}
\input{05-Experiments.tex} 
% \input{05-EntityRelationExtraction.tex}
% \input{06-OverlappingEntities.tex}
%   \input{05-ExperimentalSetup.tex}
%  \input{06-Experiments.tex}
  \input{07-Analysis.tex}
\input{08-Conclusion.tex}
\input{09-Acknowledgement.tex}
%  \input{04-GraphProp.tex}

% \clearpage

% \clearpage
% \input{04-KGConstruction.tex}
% \input{05-Experiment.tex}
% \input{05-Experiments.tex}
% \input{06-Analysis.tex}
% \input{06-Conclusion.tex}
 \bibliography{references}
\bibliographystyle{acl_natbib}

\end{document}

%% file: 01-Abstract.tex
\begin{abstract}

We introduce a general framework for several information extraction tasks that share span representations using dynamically constructed span graphs. The graphs are constructed by selecting the most confident entity spans and linking these nodes with confidence-weighted relation types and coreferences. The dynamic span graph allows coreference and relation type confidences to propagate through the graph to iteratively refine the span representations. This is unlike  previous multi-task frameworks for information extraction in which the only interaction between tasks is in the shared first-layer LSTM.  Our framework significantly outperforms the state-of-the-art on multiple information extraction tasks across multiple datasets reflecting different domains.
We further observe that the span enumeration approach is good at detecting nested span entities, with significant F1 score improvement on the ACE dataset.\footnote{Code and pre-trained models are publicly available at \url{https://github.com/luanyi/DyGIE}.}

\end{abstract}

%% file: 02-Introduction.tex
\section{Introduction}
Most Information Extraction (IE) tasks require identifying and categorizing phrase spans, some of which might be nested.  For example, entity recognition involves assigning an entity label to a phrase span. Relation Extraction (RE) involves assigning a relation type between pairs of spans. Coreference resolution groups spans referring to the same entity into one cluster. 
%%MO: shortening
Thus, we might expect that knowledge learned from one task might benefit another.
%Since the three tasks share the same span-level representations, incorporating the relationships learned by each of them into these shared representations can improve the model's performance across all tasks.

Most previous work in IE (e.g., \cite{nadeau2007survey, chan2011exploiting}) employs a pipeline approach, first  detecting entities and then using the detected entity spans for relation extraction and coreference resolution. To avoid cascading errors introduced by pipeline-style systems, 
%\dwadden{this next part is tough to understand but I don't know how to rephrase it. We can talk tomorrow.} 
recent work has focused on coupling different IE tasks as in 
%information extraction tasks using pairs of tasks such as 
joint modeling of entities and relations~\cite{miwa2016end, zhang2017end}, entities and coreferences~\cite{hajishirzi2013joint, durrett2014joint},  joint inference~\cite{Singh2013JointIO} or multi-task (entity/relation/coreference) learning~\cite{luan2018multi}. These models mostly rely on the first layer LSTM to share span representations between different tasks and are usually designed for specific domains.  

% Recent work ~\cite{li2014incremental, miwa2016end, zheng2017joint} focused on address the cascading error problem through jointly modeling entity and relation extraction. However, the information flow is still single directional, where the interaction between tasks are very limited. \luanyi{rewrite the sentence}  
In this paper, we introduce a general framework Dynamic Graph IE (\sys) for coupling multiple information extraction tasks through shared span representations which are refined leveraging contextualized information from relations and coreferences.  Our framework is effective in several domains, demonstrating a benefit from incorporating broader context learned from relation and coreference annotations.

%indicating that a multi-task learning setup can benefit from enriched span representations with a  broader context learned from relation and coreference annotations. 

\input{figures/example_figure_car.tex}

Figure~\ref{fig:example} shows an example illustrating the potential benefits of entity, relation, and coreference contexts. It is impossible to predict the entity labels for \textit{This thing} and \textit{it} from within-sentence context alone. However, the antecedent  \textit{car}  strongly suggests that these two entities have a VEH type. 
Similarly, the fact that \textit{Tom} is located at \textit{Starbucks} and \textit{Mike} has a relation to \textit{Tom} provides support for the fact that  \textit{Mike} is located at \textit{Starbucks}.   
% \luanyi{Similarly, IE system can also benefit from higher order relations. For example, a sentence typically contains multiple relations between entity mentions. RE models need to consider these pairs simultaneously to model the dependencies among them. The relation between a pair of interest can be influenced by other pairs in the same sentence.}

% are among the antecedents of \textit{ours} can strongly support  \textit{ours} being classified to a PER type. coreference links  use coreference as an auxiliary task that allows the model to incorporate cross-sentence information and models higher-order relation to incorporate broader within-sentence context.

\sys\ uses multi-task learning to identify entities, relations, and coreferences through shared span representations using dynamically constructed span graphs. The nodes in the graph are dynamically selected from a beam of highly-confident mentions, and the edges are weighted according to the confidence scores of relation types or coreferences. Unlike the multi-task method that only shares span representations from the local context~\cite{luan2018multi}, our framework leverages rich contextual span representations by propagating information through coreference and relation links. Unlike previous BIO-based entity recognition systems~\cite{collobert2008unified,lample2016neural,ma2016end} that  assign a text span to at most one entity, our framework enumerates and represents all possible spans to recognize arbitrarily overlapping entities. 

We evaluate \sys\ on several datasets spanning many domains (including news, scientific articles, and wet lab experimental protocols), achieving state-of-the-art performance across all tasks and domains and demonstrating the value of coupling related tasks to learn richer span representations. For example,  \sys\ achieves relative improvements of 5.7\% and 9.9\% over state of the art on the ACE05 entity and relation extraction tasks, and an 11.3\% relative improvement on the ACE05 overlapping entity extraction task.  

%%MO: added a phrase to contributions instead
%We will make our framework publicly available, and we invite researchers to extend and run this framework on their domains of interest.

The contributions of this paper are threefold.
1)~We introduce the dynamic span graph framework as a method to propagate global contextual information, making the code publicly available.
2)~We demonstrate that our framework significantly outperforms the state-of-the-art on joint entity and  relation detection tasks across four datasets: ACE 2004, ACE 2005, SciERC and the Wet Lab Protocol Corpus.
3)~We further show that our approach excels at detecting entities with overlapping spans, achieving an improvement of up to 8 F1 points on three benchmarks annotated with overlapped spans: ACE 2004, ACE 2005 and GENIA.

%% file: figures/example_figure_car.tex
\begin{figure}[t]
\centering
\includegraphics[width=\columnwidth, keepaspectratio, trim={9cm 14cm 10.5cm 5cm}, clip]{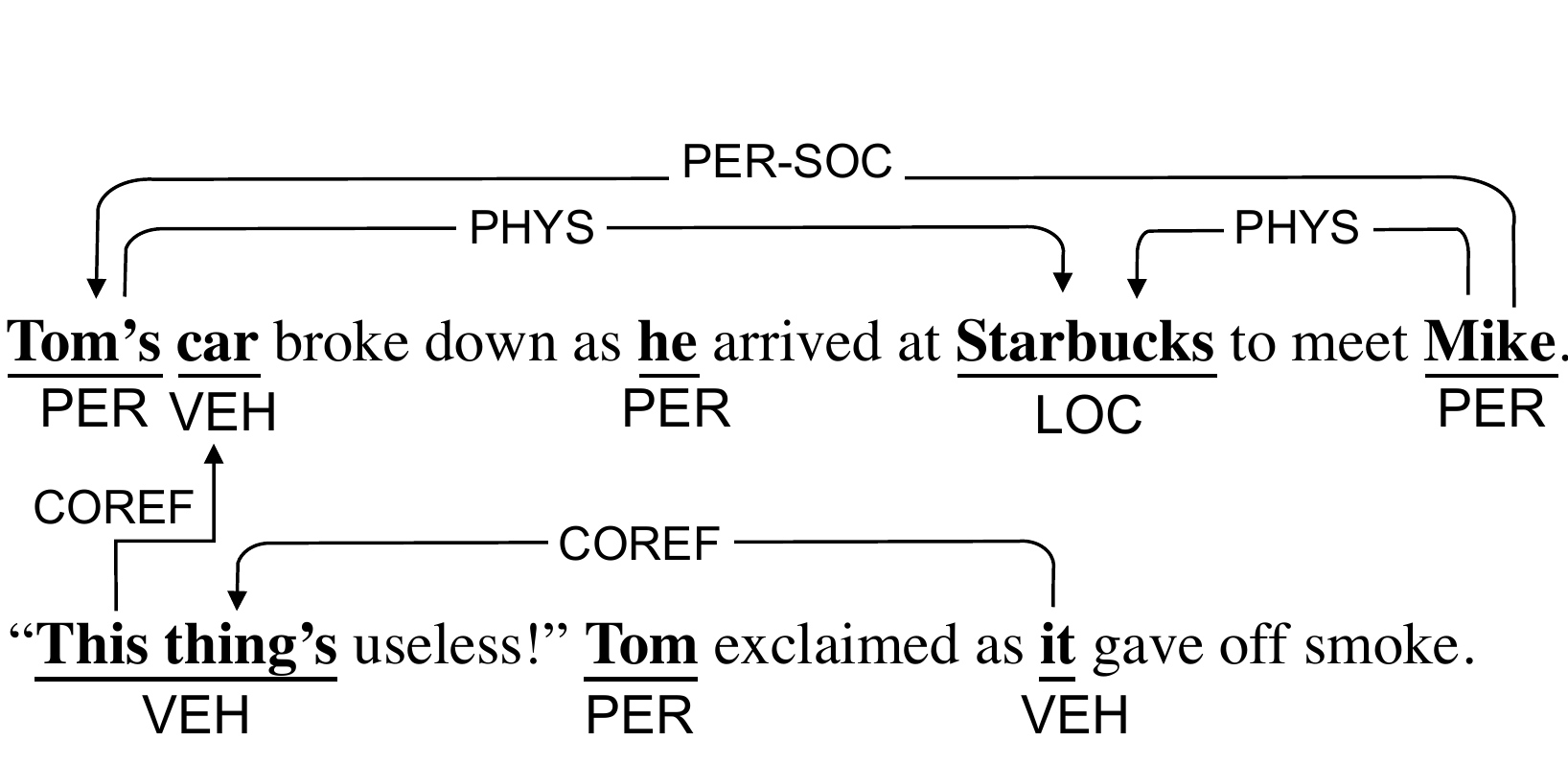}
% \vspace{-1em}
\caption{
A text passage illustrating interactions between entities, relations and coreference links. 
%Coreference links for "Tom" ommitted for clarity.
Some relation and coreference links are omitted.
}\label{fig:example}
\vspace{-1em}
\end{figure}

%% file: 03-RelatedWork.tex
\section{Related Work}

% \red{The idea of leveraging information from different tasks has started in previous work through joint modeling~\cite{miwa2016end,zhang2017end, Singh2013JointIO,yang2016joint} or multi-task learning~\cite{peng2015named,peng2017cross,luan2018multi}. \sys\ is closest to ~\newcite{luan2018multi} which is a mult-task learning framework for extracting entity, relation and coreference. Different from \newcite{luan2018multi} where the only interaction between tasks is the shared first-layer LSTM \luanyi{maybe redundant or rephrase}, \sys\ enhance the interaction through dynamic graph propagation.} 

Previous studies have explored joint modeling~\cite{miwa2016end,zhang2017end, Singh2013JointIO,yang2016joint}) and multi-task learning~\cite{peng2015named,peng2017cross,luan2018multi, luan2017multi}  as methods to share representational strength across related information extraction tasks.  The   most similar to ours is the work in~\newcite{luan2018multi} that takes a multi-task learning approach to entity, relation, and coreference extraction. In this model, the different tasks share span representations that only incorporate broader context indirectly via the gradients passed back to the LSTM layer. In contrast, \sys\ uses dynamic graph propagation to  explicitly incorporate rich contextual information into the span representations.

Entity recognition has commonly been cast as a sequence labeling problem, and has benefited substantially from the use of neural architectures ~\cite{collobert2011natural, lample2016neural, ma2016end, luan2017scienceie, luan2018uwnlp}.  However, most systems based on sequence labeling suffer from an inability to extract entities with overlapping spans. Recently \newcite{katiyar2018nested} and \newcite{Wang2018NeuralSH} have presented methods enabling neural models to extract overlapping entities, applying hypergraph-based representations on top of sequence labeling systems. Our framework  offers an alternative approach, forgoing sequence labeling entirely and simply considering all possible spans as candidate entities.

Neural graph-based models have achieved significant improvements over traditional feature-based approaches on several graph modeling tasks. Knowledge graph completion~\cite{yang2014embedding, bordes2013translating} is one prominent example.
For relation extraction tasks, graphs have been used primarily as a means to incorporate pipelined features such as  syntactic or discourse relations ~\cite{peng2017cross,song2018n, zhang2018graph}.  \newcite{christopoulou2018walk} models all possible paths between entities as a graph, and refines pair-wise embeddings by performing a walk on the graph structure. All these previous works assume that the nodes of the graph (i.e. the entity candidates to be considered during relation extraction) are predefined and fixed throughout the learning process.  On the other hand, our framework does not require a fixed set of entity boundaries as an input for graph construction. Motivated by state-of-the-art span-based approaches to coreference resolution \cite{Lee2017EndtoendNC, lee2018higher} and semantic role labeling~\cite{he2018jointly}, the model uses a beam pruning strategy to dynamically select high-quality spans, and constructs a graph using the selected spans as nodes.

 Many state-of-the-art RE models rely upon domain-specific external syntactic tools to construct dependency paths between the entities in a sentence~\cite{li2014incremental, xu2015semantic, miwa2016end, zhang2017end}. These systems suffer from cascading errors from these tools and are hard to generalize to different domains.
To make the model more general, we combine the multitask learning framework with ELMo embeddings~\cite{peters2018deep} without relying on external syntactic tools and risking the cascading errors that accompany them, and improve the interaction between tasks through dynamic graph propagation. 
While the performance of DyGIE benefits from ELMo, it advances over some systems ~\cite{luan2018multi,sanh2018hierarchical} that also incorporate ELMo. The analyses presented here give insights into the benefits of joint modeling.

%% file: 04-Model.tex
\input{figures/model_figure.tex}

\section{Model}

%\subsection{Overview}
\paragraph{Problem Definition} The input is a document represented as a sequence of words $D$, %$D=\{w_1, \ldots, w_n\}$, 
from which we derive
$S=\{s_1,\ldots, s_T\}$, the set of all possible within-sentence word sequence spans (up to length $L$) in the document.
The output contains three structures: 
the entity types $E$ for all spans $S$, 
the relations $R$ for all span pairs $S\times S$ within the same sentence, and the coreference links $C$ for all spans in $S$ across sentences.  
%  \red{In particular, {\it Entity recognition} is to predict the best entity type $e_i$ for every candidate span $s_i$. {\it Relation extraction} is to predict  the best relation type $r_{ij}$ given an ordered pair of spans $(s_i,s_j)$. {\it Coreference resolution} is to predict the best antecedent $c_i$  given a span $s_i$.}
We consider two primary tasks. First, {\it Entity Recognition} is the task of predicting the best entity type labels $e_i$ for each span $s_i$. Second, {\it Relation Extraction} involves predicting the best relation type $r_{ij}$ for all span pairs $(s_i,s_j)$.
We provide additional supervision by also training our model to perform a third, auxiliary task: {\it Coreference resolution}. For this task we predict the best antecedent $c_i$ for each span $s_i$.

\paragraph{Our Model} We develop a general information extraction framework (\sys) to identify and classify  entities, relations, and coreference in a multi-task setup.
%%MO: I took this out since Fig 2 is mentioned in the next paragraph. We don't need it twice and I prefer it in the next paragraph
%(sketched in Figure \ref{fig:model:mtl}). 
% \red{\sys\  enumerates all phrase spans in a sentence, encodes them in a vector space representation through local context, and refines the span representations using global information by propagating the entity and relation type confidences through a {\it graph of spans}. At each  training step, the graph of spans are dynamically constructed by selecting the most confident entity spans and linking these nodes with confidence-weighted relation types and coreferences.  Then, the span representations are refined using global context from  propagations of  neighboring relation types and corefered entities.}  
\sys\ first enumerates all text spans in each sentence, and computes a locally-contextualized vector space representation of each span. The model then employs a \emph{dynamic span graph} to incorporate global information into its span representations, as follows. At each training step, the model identifies the text spans that are most likely to represent entities, and treats these spans as nodes in a graph structure. It constructs confidence-weighted arcs for each node according to its predicted coreference and relation links with the other nodes in the graph.  Then, the span representations are refined using broader context from  gated updates propagated from neighboring relation types and co-referred entities. These refined span representations are used in a multi-task framework to predict entity types, relation types, and coreference links. 
% Unlike the previous multi-task framework~\cite{luan2018multi} in which the only interaction between tasks is the shared first-layer LSTM capturing local context, our model uses the global context through the span graph to  allow entity and relation type confidences to propagate through the graph, iteratively refining the \red{result} \dwadden{model's span representations}. These refined span representations are used in a multi-task framework to predict entity types, relation types, and coreference links. \mo{maybe redundant with intro}

\subsection{Model Architecture}
\label{sec:model_architecture}
%We develop a general information extraction framework (called \sys) to identify and classify scientific entities, relations, and coreference. Unlike the baseline model ~\cite{luan2018multi} in which the only interaction between tasks is the shared first-layer LSTM, the dynamic span graph allows entity and relation type confidences to propagate through the graph to iteratively refine the result. 
In this section, we give an overview of the main components and layers of the \sys\ framework, as illustrated in Figure \ref{fig:model:mtl}.
%Figure \ref{fig:model:mtl} illustrates the main components and layers of the \sys\ framework, which are overview here.
%The next section will give details of the graph construction and refinement.
Details of the graph construction and refinement process will be presented in the next section.

\paragraph{Token Representation Layer}
We apply a bidirectional LSTM over the input tokens. The input for each token is a concatenation of the character reprensetation, GLoVe~\cite{pennington2014glove} word embeddings, and ELMo embeddings \cite{peters2018deep}. The output token representations are obtained by stacking the forward and backward LSTM hidden states.

\paragraph{Span Representation Layer}
For each span $s_i$, its initial vector representation  $\mathbf{g}^0_i$ is obtained by concatenating BiLSTM outputs at  the left and right end points of $s_i$, an attention-based soft ``headword," and an embedded span width
feature, following \newcite{Lee2017EndtoendNC}.

\paragraph{Coreference Propagation Layer}
 The propagation process starts from the span representations $\mathbf{g}_i^0$. 
At each iteration $t$, we first compute an \emph{update vector} $\mathbf{u}_C^{t}$ for each span $s_i$. Then we use $\mathbf{u}_C^{t}$ to update the current representation $\mathbf{g}_i^{t}$, producing the next span representation $\mathbf{g}_i^{t+1}$.
%%MO: not needed, since details come later
%by learning the gating vector $\mathbf{f}_i^C$ according to the equation $\ref{eq:gate}$.  
By repeating this process $N$ times, the final span representations $\mathbf{g}_i^{N}$ share contextual information across spans that are likely to be antecedents in the coreference graph, similar to the process in \cite{lee2018higher}. % learn the gating function $f_i^n$ according to the equation  by using the current antecedent distribution $P(c_i = j|D)$ as an attention mechanism through Eq.~\ref{eq:link} and ~\ref{eq:gate}, where $\text{Link}_C(i, j,  \mathbf{g}_j^{n})$ is defined as:
%\begin{equation}
% \text{Link}_C(i, j,  \mathbf{g}_j^{n}, B_\text{C}) = \sum_{j\in B_\text{C}} P(c_i = j|D) \cdot \mathbf{g}_j^{n}
%\end{equation}

\paragraph{Relation Propagation Layer}
The outputs $\mathbf{g}_i^{N}$ from the coreference propagation layer are passed as inputs to the relation propagation layer.  Similar to the coreference propagation process, at each iteration $t$, we first compute the update vectors  $\mathbf{u}_R^{t}$ for each span $s_i$, then use it to compute $\mathbf{g}_i^{t+1}$.
Information can be integrated from multiple relation paths by repeating this process $M$ times. 

\paragraph{Final Prediction Layer}
We use the outputs of the relation graph layer $\mathbf{g}_i^{N+M}$ to predict the entity labels $E$ and relation labels $R$. For entities, we pass $\mathbf{g}_i^{N+M}$ to a feed-forward network (FFNN) to produce per-class scores $\mathbf{P}_E (i)$ for span $s_i$. For relations, we pass the concatenation of $\mathbf{g}_i^{N+M}$ and $\mathbf{g}_j^{N+M}$ to a FFNN to produce per-class relation scores $\mathbf{P}_R (i,j)$ between spans $s_i$ and $s_j$.
Entity and relation scores are  normalized across the label space, similar to~\newcite{luan2018multi}.
%: $P (e_i \mid C, R, \mathcal{D}) \propto \exp (\mathbf{P}_E(i) )$ and $P (r_{ij} \mid C, \mathcal{D}) \propto \exp (\mathbf{P}_R (i,j))$.\\
%
For coreference, the scores between span pairs ($s_i, s_j$) are computed from the coreference graph layer outputs ($\mathbf{g}_i^{N},\mathbf{g}_j^{N}$), and then normalized across all possible antecedents, similar to~\newcite{lee2018higher}. 
%: $P (c_i=j \mid \mathcal{D}) \propto \exp (\mathbf{P}_C (i,j))$, similar to~\newcite{Lee2017EndtoendNC}. %\luanyi{need to rephrase}

%\subsection{Span Representation Refinement through Dynamic Graph Construction}
\subsection{Dynamic Graph Construction and Span Refinement}
\label{sec:model:graph}
The dynamic span graph facilitates propagating broader contexts through soft coreference and relation links to refine span representations.
 The nodes in the graph are spans $s_i$ with vector representations $\mathbf{g}^t_i \in \mathbb{R}^{d} $ for the $t$-th iteration. The edges are weighted by the coreference and relation scores, which are trained according to the neural architecture explained in Section~\ref{sec:model_architecture}.
 In this section, we explain how coreference and relation links can update span representations.  
 
 \paragraph{Coreference Propagation}
 Similar to~\cite{luan2018multi}, we define a beam $B_C$ consisting of $b_c$ spans that are most likely to be in a coreference chain.
 We consider $\mathbf{P}_C^t$ to be a matrix of real values that indicate coreference confidence scores between these spans at the $t$-th iteration. $\mathbf{P}_C^t$ is of size $b_c \times K$, where $K$ is the maximum number of antecedents considered.
 For the coreference graph, an edge in the graph is single directional, connecting the current  span $s_i$ with all its potential antecedents $s_j$ in the coreference beam, where $j < i$.  The edge between $s_i$ and $s_j$ is weighted by coreference confidence score at the current iteration $P^t_C(i,j)$. 
The span update vector $\mathbf{u}_C^{t}(i)\in \mathbb{R}^{d}$ is computed by aggregating the neighboring span representations $\mathbf{g}_j^{t}$, weighted by their coreference scores $P_C^t(i,j)$: %are refined given the n each graph layer, we get the updated span embedding $\mathbf{a}_i^{n}$ through the following function:
\begin{equation} \label{eq:coref}
 \mathbf{u}_C^t(i)=  \sum_{j\in B_\text{C}(i)} P^t_C(i,j)\mathbf{g}_j^{t}
\end{equation}
where $B_C(i)$ is the set of $K$ spans that are antecedents of $s_i$, 
\begin{equation}
    P_C^t(i,j) = \frac{
    \exp(V_C^t(i,j))}{ \sum_{j' \in B_\text{C}(i)} \exp(V_{C}^t(i,j))}
\end{equation}
$V_C^t(i,j)$ is a scalar score computed by concatenating the span representations $[\mathbf{g}_i^t, \mathbf{g}_j^t, \mathbf{g}_i^t\odot\mathbf{g}_j^t]$, where $\odot$ is element-wise multiplication.
The concatenated vector is then fed as input to a FFNN, similar to ~\cite{lee2018higher}. %  \luanyi{Luheng:check}

\paragraph{Relation Propagation}
For each sentence, we define a beam $B_R$ consisting of $b_r$ entity spans that are mostly likely to be involved in a relation.  Unlike the coreference graph, the weights of relation edges capture different relation types.
Therefore, for the $t$-th iteration, we use a tensor $\mathbf{V}_R^{t}\in \mathbb{R}^{b_R\times b_R \times L_R} $ to capture scores of each of the $L_R$ relation types. 
In other words, each edge in the relation graph connects two entity spans $s_i$ and $s_j$ in the relation beam $B_R$. 
%The relation prediction at each iteration is a $L_R$-length vector of predicted relation 
%
$\mathbf{V}_R^t(i,j)$ is a $L_R$-length vector of relation scores, 
computed with a FFNN with $[\mathbf{g}_i^{t}, \mathbf{g}_j^{t}]$ as the input.  
%% maybe include equation or cite paper
The relation update vector $\mathbf{u}^t_{R}(i)\in \mathbb{R}^{d}$ is computed by aggregating neighboring span representations on the relation graph: 
\begin{align}\label{eq:relation}
 \mathbf{u}^t_{R}(i) =  \sum_{j\in B_\text{R}} f(\mathbf{V}^t_R(i,j))\mathbf{A}_R \odot \mathbf{g}_j^{t},
\end{align}
where $\mathbf{A}_R \in \mathbb{R}^{L_R\times d}$ is a trainable linear projection matrix, 
$f$ is a non-linear function to select the most important relations. 
Because only a small number of entities in the relation beam  are actually linked to the target span, propagation among all possible span pairs would introduce too much noise to the new representation.
Therefore, we choose $f$ to be the ReLU function to remove the effect of unlikely relations by setting the all negative relation scores to 0.
Unlike coreference connections, two spans linked via a relation are not expected to have similar representations, so the matrix $\mathbf{A}_R$ helps to transform the embedding $\textbf{g}_j^t$ according to each relation type. 
% \luheng{this is a bit hand-wavvy. let's reword it and add it back for camera-ready?}

\paragraph{Updating Span Representations with Gating} 
%\paragraph{Gated Representation Updates} 
To compute the span representations for the next iteration $t \in \{1,\ldots, N+M\}$, we define a gating vector $\mathbf{f}_x^{t}(i)\in \mathbb{R}^{d}$, where $x\in\{C,R\}$, to determine whether to keep the previous span representation $\mathbf{g}_i^{t}$ or to integrate new information from the coreference or relation update vectors $\mathbf{u}^t_{x}(i)$. Formally, 
\begin{eqnarray}
\mathbf{f}_x^t(i) &=& g(\mathbf{W}^{\text{f}}_x[\mathbf{g}_i^{t},\mathbf{u}^t_{x}(i)]) \label{eq:gate}\\
\mathbf{g}_i^{t+1} &=& \mathbf{f}_x^t(i)\odot \mathbf{g}_i^{t} + (1-\mathbf{f}_x^t(i)) \odot \mathbf{u}^t_x(i), \notag
\end{eqnarray}
% \hanna{for ui and fi, you don't need to have the superscript n, right? I am adding superscript X, where X is either R or C}
% \hanna{This shouldn't be Wx, we already have used WR.} \hanna{don't do sigma, do g, and in the implementation details or in the text say you used sigmoid}
where $\mathbf{W}^{\text{f}}_x \in \mathbb{R}^{d\times 2d}$ are trainable parameters, and $g$ is an element-wise sigmoid function.
% \hanna{The iterations in the propagations are not explained -- maybe this can be deferred to the model architecture} 

\subsection{Training}

The loss function is defined as a weighted sum of the  log-likelihood of all three tasks:
\begin{align}
    %\mathcal{J}(R^*, E^*, C^*, D) =& 
 & \sum_{(D, R^*, E^*, C^*) \in\mathcal{D}} \Big\{ 
 \lambda_{\text{E}}\log P (E^* \mid C, R, D)   \\
 & + \lambda_{\text{R}}\log P (R^* \mid C, D) + \lambda_{\text{C}}\log P (C^* \mid D) \Big\}\notag
\end{align}
where $E^*$, $R^*$ and $C^*$ are gold structures of the entity types, relations and coreference, respectively. $\mathcal{D}$ is the collection of all training documents $D$.
The task weights $\lambda_{\text{E}}$, $\lambda_{\text{R}}$, and $\lambda_{\text{C}}$ are   hyper-parameters to control the importance of each task.

 We use a 1~layer BiLSTM with 200-dimensional hidden layers. All the feed-forward functions have 2 hidden layers of 150 dimensions each.  We use 0.4 variational dropout~\cite{gal:2016} for the LSTMs, 0.4 dropout for the FFNNs, and 0.5 dropout for the input embeddings.
 The hidden layer dimensions and dropout rates are chosen based on the development set performance in multiple domains.
 The task weights, learning rate, maximum span length, number of propagation iterations and beam size are tuned specifically for each dataset using development data.
%%MO: old version
%For maximum span width and beam size, we tune the parameters based on the performance of the development set and varies across different datasets.
%We use the same batching process as \newcite{luan2018multi} that randomly splits documents to fit each batch. 
%and randomly truncate long sentences to batch size to prevent memory overflow \luanyi{Luheng Please check this}.
%We train the model using Adam. 
%\cite{We should find citation for Adam, but i'm too tired to look for it. let's add this for camera-ready.}
 

%% file: figures/model_figure.tex
\begin{figure*}[t]
\centering
\includegraphics[width=0.82\textwidth, keepaspectratio, trim={0.75cm 2cm 0.75cm 3.5cm}, clip]{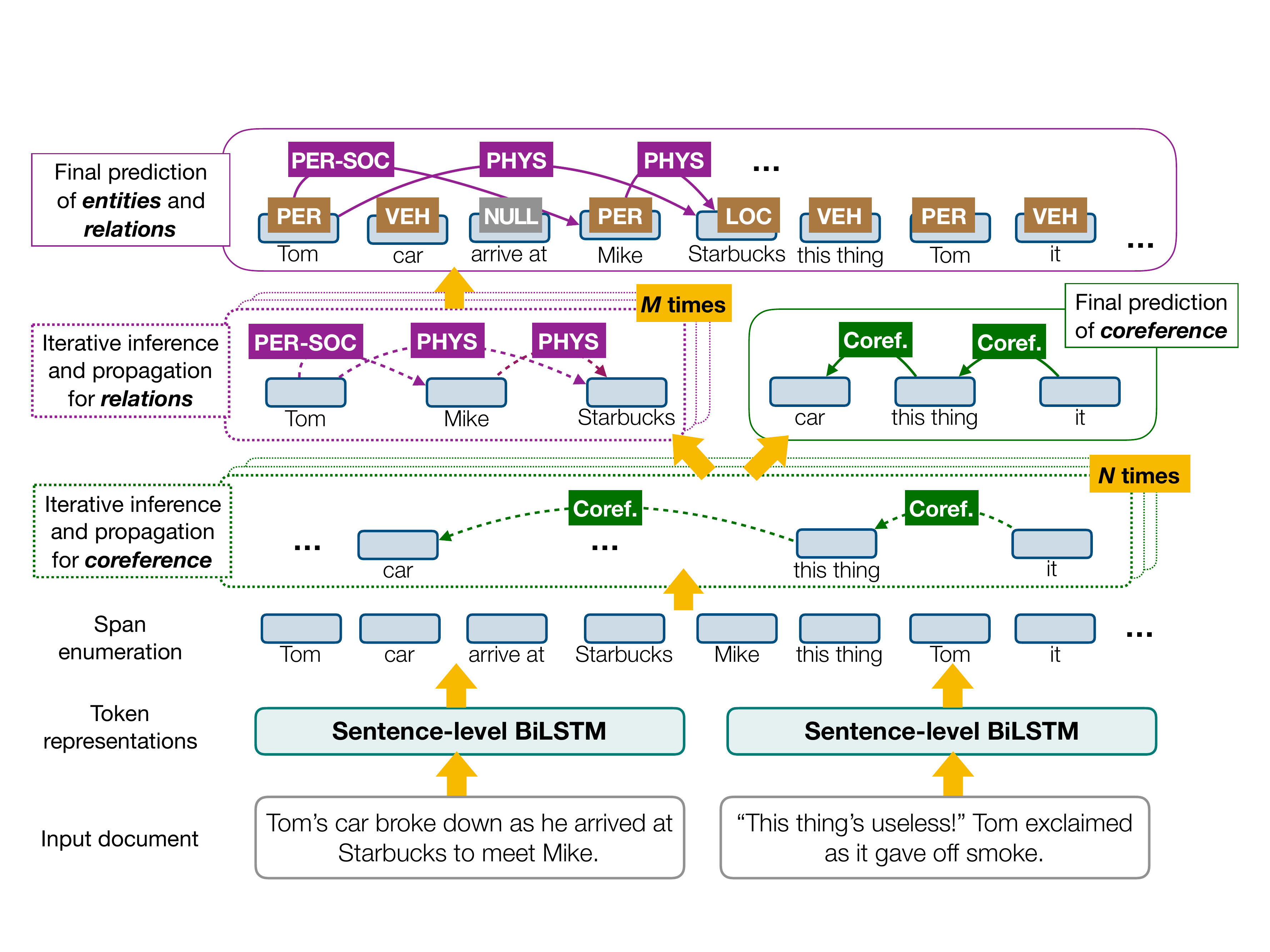}
% \vspace{-1em}
\caption{
Overview of our \sys\ model. Dotted arcs indicate confidence weighted graph edges. Solid lines indicate the final predictions.
%Information is propagated inside coreference beam and relation beam through multiple iterations.
}\label{fig:model:mtl}
\vspace{-1em}
\end{figure*}

%% file: 05-Experiments.tex
\section{Experiments}
\sys\ is a general IE framework that can be applied to multiple  tasks. We evaluate the performance of \sys\ against models from two lines of work: combined entity and relation extraction, and overlapping entity extraction.

\input{05-EntityRelationExtraction.tex}
\input{06-OverlappingEntities.tex}

%% file: 05-EntityRelationExtraction.tex
\subsection{Entity and relation extraction}
\label{sec:joint_entity_rel}

For the entity and relation extraction task, we test the performance of \sys\ on four different datasets: ACE2004, ACE2005, SciERC and the Wet Lab Protocol Corpus.
We include the relation graph propagation layer in our models for all datasets. We include the coreference graph propagation layer on the data sets that have coreference annotations available.
\paragraph{Data}

All four data sets are annotated with entity and relation labels. Only a small fraction of entities ($<3\%$ of total) in these data sets have a text span that overlaps the span of another entity.
Statistics on all four data sets are displayed in Table~\ref{tab:data_joint}.

\input{tables/data_summary_joint.tex}

The \textbf{ACE2004} and \textbf{ACE2005} corpora provide entity and relation labels for a collection of documents from a variety of domains, such as newswire and online forums. We use the same entity and relation types, data splits, and preprocessing as \newcite{miwa2016end} and \newcite{li2014incremental}. Following the convention established in this line of work, an entity prediction is considered correct if its type label and head region match those of a gold entity.  We will refer to this version of the ACE2004 and ACE2005 data as ACE04 and ACE05. Since the domain and mention span annotations in the ACE datasets are very similar to those of OntoNotes~\cite{pradhan2012conll}, and OntoNotes contains significantly more documents with coreference annotations, we use OntoNotes to train the parameters for the auxiliary coreference task. The OntoNotes corpus contains 3493 documents, averaging roughly 450 words in length.

% \luanyi{add statistics of ontonotes}
 
The \textbf{SciERC} corpus \cite{luan2018multi} provides entity, coreference and relation annotations for a collection of documents from 500 AI paper abstracts.   The dataset defines scientific term types and relation types specially designed for AI domain knowledge graph construction. An entity prediction is considered correct if its label and span match with a gold entity.

The \textbf{Wet Lab Protocol Corpus (WLPC)} provides entity, relation, and event annotations for 622 wet lab protocols~\cite{Kulkarni2018AnAC}. A wet lab protocol is a series of instructions specifying how to perform a biological experiment. Following the procedure in \newcite{Kulkarni2018AnAC}, we perform entity recognition on the union of entity tags and event trigger tags, and relation extraction on the union of entity-entity relations and entity-trigger event roles. Coreference annotations are not available for this dataset.

\input{tables/results_jointb.tex}

\paragraph{Baselines}
We compare \sys\ with current state of the art methods in different datasets. 
\newcite{miwa2016end} provide the current state of the art on ACE04. They construct a Tree LSTM using dependency parse information, and use the representations learned by the tree structure as features for relation classification. \newcite{bekoulis2018adversarial} use adversarial training as regularization for a neural model. \newcite{zhang2017end} cast joint entity and relation extraction as a table filling problem and build a globally optimized neural model incorporating syntactic representations from a dependency parser.  Similar to \sys, \newcite{sanh2018hierarchical} and \newcite{luan2018multi} use a multi-task learning framework for extracting entity, relation and coreference labels. \newcite{sanh2018hierarchical} improved the state of the art on ACE05 using multi-task, 
 hierarchical supervised training with a set of low level tasks at the bottom layers of the model and more complex tasks at the top layers of the model.  \newcite{luan2018multi} previously achieved the state of the art on SciERC and use a span-based neural model like our {\sys}.  \newcite{Kulkarni2018AnAC} provide a baseline for the WLPC data set. They employ an LSTM-CRF for entity recognition, following \newcite{lample2016neural}. For relation extraction, they assume the presence of gold entities and train a maximum-entropy classifier using features from the labeled entities.

\paragraph{Results}

Table~\ref{tab:results_joint} shows test set F1 on the joint entity and relation extraction task. We observe that \sys\ achieves substantial improvements on both entity recognition and relation extraction across the four data sets and three domains, all in the realistic setting where no ``gold'' entity labels are supplied at test time. \sys\ achieves 7.1\% and 7.0\% relative improvements over the state of the art on NER for ACE04 and ACE05, respectively.  For the relation extraction task, \sys\ attains 25.8\% relative improvement over SOTA on ACE04 and 13.7\% relative improvement on ACE05. For ACE05, the best entity extraction performance is obtained by switching the order between $\texttt{CorefProp}$ and $\texttt{RelProp}$ ($\texttt{RelProp}$ first then $\texttt{CorefProp}$). 

On SciERC, \sys\ advances the state of the art by 5.9\% and 1.9\% for relation extraction and NER, respectively. The improvement of \sys\ over the previous SciERC model underscores the ability of coreference and relation propagation to construct rich contextualized representations.

The results from \newcite{Kulkarni2018AnAC} establish a baseline for IE on the WLPC. In that work, relation extraction is performed using gold entity boundaries as input. Without using any gold entity information, \sys\ improves on the baselines by 16.8\% for relation extraction and 2.2\% for NER.

On the OntoNotes data set used for the auxiliary coreference task with ACE05, our model achieves coreference test set performance of 70.4 F1, which is competitive with the state-of-the-art performance reported in ~\newcite{Lee2017EndtoendNC}.

%% file: tables/data_summary_joint.tex
\begin{table}[t]
\centering
\footnotesize
\begin{tabular}{L{1.5cm}L{1.0cm}R{0.4cm}R{0.4cm}R{0.4cm}C{0.6cm}}
\toprule
& Domain & Docs & Ent & Rel & Coref\\
\midrule
ACE04 & News & 348 & 7 & 7 & \cmark\\
ACE05 & News & 511 & 7 & 6 & \xmark\\
SciERC & AI & 500 & 6 & 7 & \cmark\\
WLP & Bio lab & 622 & 18 & 13 & \xmark\\
\bottomrule
\end{tabular}
\caption{Datasets for joint entity and relation extraction and their statistics. \emph{Ent}: Number of entity categories. \emph{Rel}: Number of relation categories.}
\label{tab:data_joint}
\vspace{-1em}
\end{table}

%% file: tables/results_jointb.tex
\begin{table}[t]
\newcolumntype{Y}{>{\centering\arraybackslash}X}
\newcommand{\colindent}{\;}

\footnotesize
\centering

\begin{tabularx}{\columnwidth}{l l *{2}{Y}}
\toprule
%& & \multicolumn{2}{c}{System}  \\ % & \multicolumn{2}{c}{\sys} & \multicolumn{2}{c}{Absolute change} & \multicolumn{2}{c}{Relative change} \\
%\cmidrule(lr){3-4} % \cmidrule(lr){5-6} \cmidrule(lr){7-8} \cmidrule(lr){9-10}
Dataset & System & Entity & Relation \\% & NER & REL  & NER & REL & NER & REL \\    
\midrule
% & \textbf{87.4} & \textbf{60.9} & 5.6 & 12.5 & 6.8 & 25.8 \\
\multirow{3}{*}{ACE04} & \newcite{bekoulis2018adversarial} & 81.6 & 47.5 \\ % & - & - & - & - & - & - \\
  & \newcite{miwa2016end} & 81.8 & 48.4 \\ 
 & \sys & \textbf{87.4} & \textbf{59.7} \\
\cmidrule(lr){1-4}
% & \textbf{88.4} & \textbf{63.2} & 4.8 & 5.7 & 5.7 & 9.9 \\
 \multirow{4}{*}{ACE05} & \newcite{miwa2016end} & 83.4 & 55.6 \\ % & - & -  & - & - & - & - \\
 & \newcite{zhang2017end} & 83.6 & 57.5 \\
 & \newcite{sanh2018hierarchical} & 87.5 & 62.7 \\
 & \sys & \textbf{88.4} & \textbf{63.2} \\
\cmidrule(lr){1-4}
\multirow{2}{*}{SciERC}  & \newcite{luan2018multi}  & 64.2 & 39.3 \\ 
% & \textbf{65.4} & \textbf{41.6} \\ & 1.2 & 2.3 & 1.9 & 5.9 \\
& \sys & \textbf{65.2} & \textbf{41.6} \\
\cmidrule(lr){1-4}
\multirow{2}{*}{WLPC}    & \newcite{Kulkarni2018AnAC} & 78.0 & *54.9 \\
% & \textbf{79.7}  & *64.1 & 1.7 & *9.2 & 2.7 & *16.7 \\
& \sys &  \textbf{79.5}  & \textbf{64.1} \\
\bottomrule
\end{tabularx}
\caption{F1 scores on the joint entity and relation extraction task on each test set, compared against the previous best systems. * indicates relation extraction system that takes gold entity boundary as input.}

\label{tab:results_joint}
\vspace{-1em}

\end{table}

%% file: 06-OverlappingEntities.tex
\subsection{Overlapping Entity Extraction}
There are many applications where the correct identification of overlapping entities is crucial for correct document understanding. For instance, in the biomedical domain, a \emph{BRCA1 mutation carrier} could refer to a patient taking part in a clinical trial, while \emph{BRCA1} is the name of a gene. 

We evaluate the performance of \sys\ on overlapping entity extraction in three datasets: ACE2004, ACE2005 and GENIA. Since relation annotations are not available for these datasets, we include the coreference propagation layer in our models but not the relation layer.\footnote{We use the pre-processed ACE dataset from previous work and relation annotation is not available.}

\paragraph{Data}

Statistics on our three datasets are listed in Table~\ref{tab:data_overlap}. All three have a substantial number ($>20\%$ of total) of overlapping entities, making them appropriate for this task.

\input{tables/data_summary_overlap.tex}

As in the joint case, we evaluate our model on \textbf{ACE2004} and \textbf{ACE2005}, but here we follow the same data preprocessing and evaluation scheme as \newcite{Wang2018NeuralSH}. We refer to these data sets as ACE04-O and ACE05-O. Unlike the joint entity and relation task in Sec.~\ref{sec:joint_entity_rel}, where only the entity head span need be predicted, an entity prediction is considered correct in these experiments if both its entity label and its full text span match a gold prediction. This is a more stringent evaluation criterion than the one used in Section \ref{sec:joint_entity_rel}. As before, we use the OntoNotes annotations to train the parameters of the coreference layer.

The \textbf{GENIA} corpus~\cite{Kim2003GENIAC} provides entity tags and coreferences for 1999 abstracts from the biomedical research literature. We only use the IDENT label to extract coreference clusters. We  use the same data set split and preprocessing procedure as \newcite{Wang2018NeuralSH} for  overlapping entity recognition.

\paragraph{Baselines}

The current state-of-the-art approach on all three data sets is \newcite{Wang2018NeuralSH}, which uses a segmental hypergraph coupled with neural networks for feature learning. \newcite{katiyar2018nested} also propose a hypergraph approach using a recurrent neural network as a feature extractor.

\paragraph{Results}

\input{tables/results_overlapb.tex}

% \hanna{is the task called, mention extraction? or identification? in the literature?} \dwadden{In Wang et al they call it "overlapping mention recognition". In Katiyar and Cardie they call it "Nested named entity recognition" Why don't we refer to it as "overlapping entity recognition", since we call the other task "joint entity and relation extraction", not "joint mention and relation extraction"?}

Table~\ref{tab:results_overlap} presents the results of our overlapping entity extraction experiments on the different datsets.  \sys\ improves 11.6\% on the state of the art for ACE04-O and 11.3\% for ACE05-O. \sys\ also advances the state of the art on GENIA, albeit by a more modest 1.5\%. Together these results suggest that \sys\ can be utilized fruitfully for information extraction across different domains with overlapped entities, such as bio-medicine.

%% file: tables/data_summary_overlap.tex
% Old version.

% \begin{table*}[h]
% \centering
% {\footnotesize
% \begin{tabular}{lllrrr}
% \toprule
% & Domain & Split from & ner labels & rel labels & corpus size \\
% \midrule
% ACE04-J & newswire & \cite{li2014incremental} & 7 & 7 & 348 \\
% ACE04-O & newswire & \cite{Lu2015JointME} & 7 & 7 & 443 \\
% ACE05-J & newswire & \cite{li2014incremental} & 7 & 6 & 511 \\
% ACE05-O & newswire & \cite{Lu2015JointME} & 7 & 6 & 437 \\
% GENIA & biomedical & \cite{Wang2018NeuralSH} & 5 & - & 1999 \\
% Wet Lab Protocols & experimental process & \cite{Kulkarni2018AnAC} & 18 & 13 & 622 \\
% \bottomrule
% \end{tabular}
% }
% \caption{Summary of data sets \luanyi{split into 2 single column tables (joint entity relation v.s. overlapping mention), maybe also add number of sentences, mentions and relations if space allows.} \luanyi{where is SciERC?} \luanyi{we can probably remove ner labels and rel labels and say that in text} \luanyi{remove ``split from" column}}

% \label{tab:data}
% \end{table*}

\begin{table}[t]

\centering
\footnotesize
\begin{tabular}{L{1.3cm}L{1.0cm}R{0.5cm}C{0.4cm}C{0.9cm}C{0.6cm}}
\toprule
& Domain & Docs & Ent & Overlap & Coref\\
\midrule
ACE04-O & News & 443 & 7 & 42\% & \cmark\\
ACE05-O & News & 437 & 7 & 32\% & \xmark\\
GENIA & Biomed & 1999 & 5 & 24\% & \cmark\\
\bottomrule

% Got the percentages from http://www.statnlp.org/research/ie/emnlp2017-mention-separators.pdf

\end{tabular}
\caption{Datasets for overlapping entity extraction and their statistics. \emph{Ent}: Number of entity categories. \emph{Overlap}: Percentage of sentences that contain overlapping entities.}
\label{tab:data_overlap}
\vspace{-1em}

\end{table}

%% file: tables/results_overlapb.tex
\begin{table}[t]
\newcolumntype{Y}{>{\centering\arraybackslash}X}
\newcommand{\colindent}{\;}

\footnotesize
\centering

\begin{tabularx}{\columnwidth}{l l *{1}{Y}}
\toprule
% & & \multicolumn{1}{c}{Prev.} & \sys & Abs. change & Rel. change \\
% \cmidrule(lr){2-3} \cmidrule(lr){4-4} \cmidrule(lr){5-5} \cmidrule(lr){6-6}
% & & NER & NER & NER & NER \\    
Dataset & System & Entity F1 \\
\midrule
\multirow{3}{*}{ACE04-O} & \newcite{katiyar2018nested} & 72.7 \\ % & 83.8  & 8.7 & 11.6 \\
 & \newcite{Wang2018NeuralSH} & 75.1\\
 & \sys & \textbf{84.7}  \\
\cmidrule(lr){1-3} 
\multirow{3}{*}{ACE05-O} & \newcite{katiyar2018nested} & 70.5 \\ % & 82.9 % & 8.4 & 11.3 \\
 &  \newcite{Wang2018NeuralSH} & 74.5 \\
 & \sys & \textbf{82.9} \\
\cmidrule(lr){1-3}  
\multirow{3}{*}{GENIA}  & \newcite{katiyar2018nested}  & 73.8 \\ %  76.2  % & 1.1 & 1.5 \\
  & \newcite{Wang2018NeuralSH}  & 75.1  \\
  & \sys &  \textbf{76.2} \\
\bottomrule
\end{tabularx}
\caption{Performance on the overlapping entity extraction task, compared to previous best systems. 
We report F1 of extracted entities on the test sets.
} \label{tab:results_overlap}
\vspace{-1em}
\end{table}

%% file: 07-Analysis.tex
% \section{Coreference}

% Most of the datasets have coreference annotations included, which we use for constructing the graph layer. For the general domain datasets that does not have coreference annotations, we use OntoNotes coreference annotation as an auxiliary task to train the parameters in the coreference graph layer. \luheng{Did you mean combining OntoNotes data with the task training data, or?} For other domains do not have coreference annotation, we remove the coreference layer.  We  treat coference as an auxiliary task and not aiming at improve the performace of coreference. \luanyi{maybe too much detail about coref? do we need to report coref scores?}

\section{Analysis of Graph Propagation}

\input{tables/ablation_ace.tex}
\input{tables/ablation_scierc.tex}
\input{figures/iteration_figure.tex}
We use the dev sets of ACE2005 and SciERC to analyze the effect of different model components.

\subsection{Coreference and Relation Graph Layers}
\label{sec:analysis:graph}
Tables~\ref{tab:ace_ablation:layer} and~\ref{tab:scierc_ablation:layer} show the  effects of  graph propagation on entity and relation prediction accuracy, 
%over ACE05 and SciERC development set respectively,
where $-\texttt{CorefProp}$  and $-\texttt{RelProp}$ denote ablating the propagation process by setting $N = 0$ or $M = 0$, respectively.
$\texttt{Base}$ is the base model without any propagation.
%%MO: we can add this if the paper is accepted, for now leave it out to make the review blind
%, equivalent to \newcite{luan2018multi}. \luanyi{do we want to explicitely say equivalent?}
%with both $N = 1$ and $M = 1$ which is equal to a multi-task learning model with no propagating iterations. \luanyi{rewording} 
For ACE05, we observe that coreference propagation is mainly helpful for entities;
it appears to hurt relation extraction.
On SciIE, coreference propagation gives a small benefit on both tasks.
Relation propagation significantly benefits both entity and relation extraction in both domains.
%has more performance gain on entity extraction  in ACE05 by 1.0 F1 score, which proves that capturing document level context through coreference propagation can benefit entity extraction task. 
In particular, there are a large portion of sentences with multiple relation instances across different entities in both ACE05 and SciERC, which is the scenario in which we expect relation propagation to help.

%%MO: I moved all the pronoun stuff to the next section

%For relation extraction tasks, adding the relation layer significantly benefits both datasets, with 2.6 F1 score improvement on ACE05 and 1.7 F1 score improvement on SciERC. This shows that capturing broader context through relation propagation can benefit relation extraction task. There are a large portion of sentences with multiple relation instances across difference entities in both ACE05 and SciERC, so the improvement of relation extraction on both datasets is significant when adding relation propagation.

Since coreference propagation has more effect on entity extraction and relation propagation has more effect on relation extraction, we mainly focus on ablating the effect of coreference propagation on entity extraction and relation propagation on relation extraction in the following subsections.

\subsection{Coreference Propagation and Entities} \label{sec:analysis:document_context}

A major challenge of ACE05 is to disambiguate the entity class for pronominal mentions, which requires reasoning with cross-sentence contexts. 
For example, in a sentence from ACE05 dataset, ``One of \textbf{[them]$\blue{\mathrm{_{\bm{PER}}}}$}, from a very close friend of   \textbf{[ours]$\blue{\mathrm{_{\bm{ORG}}}}$}." It is impossible to identity whether \textit{them} and \textit{ours} is a person (\textit{PER}) or organization (\textit{ORG}) unless we have read  previous sentences.  
We hypothesize that this is a context where coreference propagation can help. 
Table~\ref{tab:pronoun_ablation} shows the effect of the coreference layer for entity categorization of pronouns.\footnote{Pronouns included: \texttt{anyone, everyone, it, itself, one, our, ours, their, theirs, them, themselves, they, us, we, who}} \sys\ has 6.6\% improvement on pronoun performance, confirming our hypothesis. 

 %Though, of course, there's alywas the chance that these fund managers were stupid and shortsighted

%\luanyi{newly added} We further include the confusion matrix between \textit{PER}, \textit{ORG} and \textit{GPE}, the most popular entity types for pronouns. Table~\ref{tab:confusion_matrix}  shows that \sys\ significantly reduces the confusion between \textit{PER} and \textit{ORG} as well as  \textit{PER} and \textit{GPE}. 

Looking further, Table~\ref{tab:confusion_matrix}  shows the impact on all entity categories, giving the difference between the confusion matrix entries with and without \texttt{CorefProp}. The frequent confusions associated with pronouns (\textit{GPE/PER} and \textit{PER/ORG}, where \textit{GPE} is a geopolitical entity) greatly improve, but the benefit of \texttt{CorefProp} extends to most categories. 

%\luanyi{add error example here} 
Of course, there are a few instances where \texttt{CorefProp} causes errors in entity extraction. For example, in the sentence ``[They]$\mathrm{_{\blue{\bm{PER}}}^{\red{\bm{ORG}}}}$
 might have been using Northshore...", \sys\ predicted \textit{They} to be of \textit{ORG} type because the most confident antecedent  is \textit{those companies} in the previous sentence: ``The money was invested in \textit{those companies}." However, \textit{They} is actually referring to \textit{these fund managers} earlier in the document, which belongs to \textit{PER} category.

In the SciERC dataset, the pronouns are uniformly assigned with a \textit{Generic} label, 
which explains why \texttt{CorefProp} does not have much effect on entity extraction performance.

% In order to prove that our coreference layer can efficiently incorporate global context and help disambiguate the entity classes, we report the confusion matrix in Table~\ref{tab:confusion_matrix} without and with coreference on the most ambiguite entity types  mostly for pron/ouns: \textit{PER}, \textit{ORG} and \textit{GPE}.  We could observe that introducing coreference can significantly reduce the confusion between \textit{PER} with \textit{ORG} (from 80 to 54) and \textit{GPE} (from 91 to 65). Moreover, the number of correctly identified classes for \textit{PER}, \textit{ORG} and \textit{GPE} significantly increased with coreference layer introduced. 

The Figure~\ref{fig:ablation:iterationa} shows the effect of number of iterations for coreference propagation in the entity extraction task. The figure shows that coreference layer obtains the best performance on the second iteration ($N=2$).

% In order to prove that our coreference layer can efficiently incorporate global context and help disambiguate the entity classes, we report the confusion matrix in Table~\ref{tab:confusion_matrix} without and with coreference on the most ambiguite entity types  mostly for pron/ouns: \textit{PER}, \textit{ORG} and \textit{GPE}.  We could observe that introducing coreference can significantly reduce the confusion between \textit{PER} with \textit{ORG} (from 80 to 54) and \textit{GPE} (from 91 to 65). Moreover, the number of correctly identified classes for \textit{PER}, \textit{ORG} and \textit{GPE} significantly increased with coreference layer introduced. 
% \luanyi{add pronoun result here}. 

\input{tables/ablation_pronoun.tex}

\input{tables/confusion_matrix_full.tex}

\subsection{Relation Propagation Impact}
\input{figures/entity_breakdown.tex}
Figure~\ref{fig:entity_breakdown} shows relation scores as a function of number of entities in sentence for \sys\ and \sys\ without relation propagation on ACE05.
The figure indicates that relation propagation achieves significant improvement in sentences with more entities, where one might expect that using broader context could have more impact.
%hinting that \sys\ is able to incorporate relation context from other entities in the sentence. 

 Figure~\ref{fig:ablation:iterationb} shows the effect of number of iterations for relation propagation in the relation extraction task. Our model achieves   the best performance on the second iteration ($M=2$).

% \input{tables/confusion_matrix.tex}
% \subsection{Overlapping entity extraction}

% % \hanna{Again, flip row and columns; It is good to bring in SoTA}
% \input{tables/overlap_pr.tex}

% To examine the ability of \sys\ to retrieve overlapping entities, we evaluate model predictions separately on sentences with at least one overlap, and sentences with no overlaps. The results are in Table \ref{tab:overlap_pr}. In both portions,
% \sys\ achieves significant improvements, with 14.3\% relative improvement on non-overlapping entities and 7.0\% improvement on overlapping entities. 

%\input{tables/ablation_num_entities.tex}

%% file: tables/ablation_ace.tex
\begin{table}[t]
\newcolumntype{Y}{>{\centering\arraybackslash}X}
\newcommand{\colindent}{\;}
\footnotesize
\centering

\begin{tabularx}{\linewidth}{l c *{6}{Y}}
\toprule
& \multicolumn{3}{c}{Entity} & \multicolumn{3}{c}{Relation} \\
\cmidrule(lr){2-4} \cmidrule(lr){5-7} 
Model & P & R & F1 & P & R & F1 \\    
\midrule
\sys\ & 87.4 & 86.7 & \textbf{87.1} & 56.2 & 60.9 & 58.4\\
\;\; $-\texttt{CorefProp}$ & 86.2 & 85.2 & 85.7 & 64.3 & 56.7 & \textbf{60.2} \\
\;\; $-\texttt{RelProp}$ & 87.0 & 86.7 & 86.9 & 60.4 & 55.8 & 58.0 \\
\;\; $\texttt{Base}$     & 86.1 & 85.7 & 85.9  & 59.5 & 55.7&57.6  \\
\bottomrule
\end{tabularx}
\caption{Ablations on the ACE05 development set with different graph propagation setups. $-\texttt{CorefProp}$ ablates the coreference propagation layers, while $-\texttt{RelProp}$ ablates the relation propagation layers.
\texttt{Base} is the system without any propagation. }\label{tab:ace_ablation:layer}
\vspace{-1em}
\end{table}

%% file: tables/ablation_scierc.tex
\begin{table}[t]
\newcolumntype{Y}{>{\centering\arraybackslash}X}
\newcommand{\colindent}{\;}
\footnotesize
\centering

\begin{tabularx}{\linewidth}{l c *{6}{Y}}
\toprule
& \multicolumn{3}{c}{Entity} & \multicolumn{3}{c}{Relation} \\
\cmidrule(lr){2-4} \cmidrule(lr){5-7} 
Model & P & R & F1 & P & R & F1 \\    
\midrule
\sys\ & 68.6 & 67.8 & \textbf{68.2} & 46.2 & 38.5 & \textbf{42.0}\\
\;\; $-\texttt{CorefProp}$ & 69.2 & 66.9 & 68.0 & 42.0 & 40.5 & 41.2 \\
\;\; $-\texttt{RelProp}$ & 69.1 & 66.0 & 67.5 & 43.6 & 37.6 & 40.4 \\
\;\; \texttt{Base}    & 70.0 & 66.3 & 68.1  &  45.4 & 34.9 & 39.5  \\
\bottomrule
\end{tabularx}
\caption{Ablations on the SciERC development set on different graph progation setups. \texttt{CorefProp} has a much smaller effect on entity F1 compared to ACE05. }\label{tab:scierc_ablation:layer}
\vspace{-1em}
\end{table}

%% file: figures/iteration_figure.tex
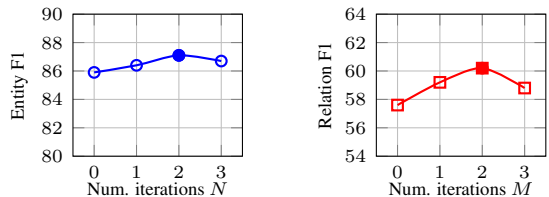
\begin{figure}[t]
\centering
\begin{subfigure}[b]{0.45\columnwidth}
\hspace{-.5em}
\begin{tikzpicture}
\begin{axis}[
    	width=1.1\columnwidth,
	    height=\columnwidth,
	    legend style={at={(1, 1)},anchor=north east,font=\scriptsize},
	    mark options={mark size=2},
		font=\scriptsize,
		xlabel near ticks,
		ylabel near ticks,
	    xmin=-0.5,xmax=3.5,
   		ymin=80, ymax=90,
   	 	ymajorgrids=true,
   	 	ytick distance=2,
    	xmajorgrids=true,
    	xlabel style={yshift=1.2ex,},
    	xlabel=Num. iterations $N$,
        ylabel=Entity F1,
    	ylabel style={yshift=-0.5ex,},
    	 title style={at={(0.5,0.92)},anchor=south,xshift=-0.5},
    	 %title=\sys$-\texttt{RelProp}$,
    	]
    \addplot[smooth,mark=o,blue,thick] plot coordinates {
    (0,85.9)
(1,86.4)
(2,87.1)
(3,86.7)
    };
    \addplot[smooth,mark=*,blue,thick] plot coordinates {
(2,87.1)
    };
\end{axis}
\end{tikzpicture}
  \caption{Entity F1 with different number of \texttt{CorefProp} iterations $N$. \label{fig:ablation:iterationa}}
\end{subfigure}
\hspace{1em}
\begin{subfigure}[b]{0.45\columnwidth}
\hspace{-.5em}
\begin{tikzpicture}
\begin{axis}[
    	width=1.1\columnwidth,
	    height=1\columnwidth,
	    legend style={at={(1, 1)},anchor=north east,font=\scriptsize},
	    mark options={mark size=2},
		font=\scriptsize,
		xlabel near ticks,
		ylabel near ticks,
	    xmin=-0.5,xmax=3.5,
   		ymin=54, ymax=64,
   	 	ymajorgrids=true,
    	xmajorgrids=true,
    	ytick distance=2,
    	xlabel=Num. iterations $M$,
    	xlabel style={yshift=1.2ex,},
        ylabel=Relation F1,
    	ylabel style={yshift=-0.5ex,},
    	title style={at={(0.5,0.92)},anchor=south,xshift=-0.5},
    	 %title=\sys$-\texttt{CorefProp}$,
    	]    
    \addplot[smooth,mark=square,red,thick] plot coordinates {
(0,57.6)
(1,59.2)
(2,60.2)
(3,58.8)
    };
  \addplot[smooth,mark=*,red,thick] plot coordinates {
(2,60.2)
    };
\end{axis}
\end{tikzpicture}
\caption{Relation F1 with different number of \texttt{RelProp} iterations $M$. \label{fig:ablation:iterationb}}
\end{subfigure}
\vspace{-.5em}
\caption{F1 score of each layer on ACE development set for different number of iterations. $N=0$ or $M=0$ indicates no propagation is made for the layer. %\luanyi{coref on NER and relation on relation, confusing, not sure how to report}
\label{fig:ablation:iteration}}
\vspace{-1em}
\end{figure}

%% file: tables/ablation_pronoun.tex
\begin{table}[t]
\footnotesize
\centering
\begin{tabular}{l c c c}
\toprule
Entity Perf. on Pronouns & P & R & F1 \\
\midrule
\sys & 79.0 & 77.1 & \textbf{78.0} \\
\sys$-\texttt{CorefProp}$ & 73.8 & 72.6 & 73.2 \\
\bottomrule
\end{tabular}
\caption{Entity extraction performance on pronouns in ACE05. \texttt{CorefProp} significantly increases entity extraction F1 on hard-to-disambiguate pronouns by allowing the model to leverage cross-sentence contexts.} 
\label{tab:pronoun_ablation}
%\vspace{-1em}
\end{table}

%% file: tables/confusion_matrix_full.tex
\begin{table}[t]
% \newcolumntype{Y}{>{\centering\arraybackslash}X}
% \newcommand{\colindent}{\;}
%\setlength{\tabcolsep}{.25em}
\footnotesize
% \centering
\resizebox{0.95\columnwidth}{!}{%
\begin{tabularx}{.5\textwidth}{c|c c c c c c c}
 & LOC & WEA & GPE & PER & FAC & ORG & VEH  \\
\hhline{--------}
LOC & 5 \cellcolor[gray]{.8}& 0 & -2 & -1 & 2 & -1 & 0 \\
WEA & 0 & 3 \cellcolor[gray]{.8}& 0 & 0 & 1 & -3 & -1 \\
GPE & -3 & 0 & 31 \cellcolor[gray]{.8} & \textbf{-26} & 3 & -7 & 0 \\
PER & 0 & -2 & -3 & 18 \cellcolor[gray]{.8} & -1 & \textbf{-26} & 4 \\
FAC & 4 & -1 & 2 & -3 & 2 \cellcolor[gray]{.8} & -5 & 1 \\
ORG & 0 & 0 & 0 & -8 & -1 & 6 \cellcolor[gray]{.8} & 0 \\
VEH & 0 & -2 & -1 & 2 & 5 & -1 & 1 \cellcolor[gray]{.8} \\
\hhline{~-------}
\end{tabularx}
}
%\caption{Relative change of confusion matrix for ACE entity extraction task before and after adding \texttt{CorefProp}.  \luanyi{newly added}} 
%%MO: revised caption
\caption{Difference in the confusion matrix counts for ACE05 entity extraction associated with adding \texttt{CorefProp}.  } 
\label{tab:confusion_matrix}

\end{table}

%% file: figures/entity_breakdown.tex
\begin{figure}[t]
\begin{tikzpicture}
\begin{axis}[
    	width=1.0\columnwidth,
	    height=0.52\columnwidth,
	    legend style={at={(1, 1)},anchor=north east,font=\scriptsize},
	    mark options={mark size=2},
		font=\scriptsize,
		xlabel near ticks,
		ylabel near ticks,
	    xmin=-1.2,xmax=3.5,
   		ymin=48, ymax=72,
   	 	ymajorgrids=true,
   	 	%ytick distance=5,
    	xmajorgrids=true,
    	xlabel=Num. entities in sentence,
        xticklabel style={align=center},
	    xtick=      {-1, 0, 1,  2, 3},
	    xticklabels={2,  3, 4-5, 6-11, 12-max},
        ylabel=Relation F1,
    	ylabel style={yshift=-0.5ex,},
    	% title style={at={(0.5,0.92)},anchor=south,xshift=-0.5},
    	% title=Coreference on NER,
    	]
\addplot[mark=o,blue,thick] plot coordinates {    
    (-1,67.9)
(0,62.6)
(1,56.7)
(2,62.2)
(3,56.3)
    };
    \addlegendentry{\sys}    	
\addplot[mark=square,red,thick] plot coordinates {
    (-1,67.2)
(0,61.8)
(1,53.4)
(2,60.3)
(3,50.0)
    };
    \addlegendentry{\sys$-\texttt{RelProp}$}
\end{axis}
\end{tikzpicture}
\vspace{-.5em}
\caption{Relation F1 broken down by number of entities in each sentence. The performance of relation extraction degrades on sentences containing more entities. Adding relation propagation alleviates this problem.
\label{fig:entity_breakdown}
}
\vspace{-1em}
\end{figure}
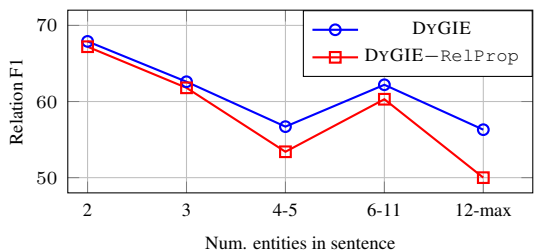

%% file: 08-Conclusion.tex
\section{Conclusion}

We have introduced \sys\ as a general information extraction framework, and have demonstrated that our system achieves state-of-the art results on entity recognition and relation extraction tasks across a diverse range of domains. The key contribution of our model is the dynamic span graph approach, which enhance interaction across tasks that allows the model to learn useful information from broader context. Unlike many IE frameworks, our model does not require any preprocessing using syntactic tools, and has significant improvement across different IE tasks including entity, relation extraction and overlapping entity extraction. 
The addition of co-reference and relation propagation across sentences adds only a small computation cost to inference; the memory cost is controlled by beam search. These added costs are small relative to those of the baseline span-based model. We welcome the community to test our model on different information extraction tasks. 
Future directions include extending the framework to encompass more structural IE tasks such as event extraction. 

% \sys\ can also be extended for other IE tasks such as event extraction, but is not investigated here and remains as our future work. \luanyi{maybe remove event?}

%% file: 09-Acknowledgement.tex
\subsection*{Acknowledgments}
This research was supported by the Office of Naval Research under the MURI grant N00014-18-1-2670, NSF (IIS 1616112, III 1703166), Allen Distinguished Investigator Award, Samsung GRO and gifts from Allen Institute for AI, Google, Amazon, and Bloomberg. We also thank the anonymous reviewers and the UW-NLP group for their helpful comments.

% This research was supported by the NSF (IIS 1616112), Allen Institute for AI (66-9175), Allen Distinguished Investigator Award, and gifts from Google, Samsung, and Bloomberg. We thank the anonymous reviewers for their helpful comments.